\documentclass[runningheads]{llncs}
\usepackage{graphicx}
\usepackage{tikz}
\usepackage{comment}
\usepackage{amsmath,amssymb} % define this before the line numbering.
\usepackage{color}

\usepackage[accsupp]{axessibility}  % Improves PDF readability for those with disabilities.

\usepackage{orcidlink}
\usepackage{graphicx}
\usepackage{amsmath,amssymb,mathabx}
\usepackage{booktabs}
\usepackage{algorithmic}
\usepackage[linesnumbered,ruled,vlined]{algorithm2e}
\usepackage{acronym}
\usepackage{enumitem}
\usepackage[capitalise]{cleveref}
\usepackage{tabularx,colortbl,multirow,array,makecell}
\usepackage{overpic,wrapfig}

\makeatletter
\DeclareRobustCommand\onedot{\futurelet\@let@token\@onedot}
\def\@onedot{\ifx\@let@token.\else.\null\fi\xspace}
\def\eg{\emph{e.g}\onedot}

\def\ie{\emph{i.e}\onedot}

\def\wrt{w.r.t\onedot}

\def\etal{\emph{et al}\onedot}
\makeatother

\DeclareMathOperator*{\argmax}{arg\,max}
\DeclareMathOperator*{\argmin}{arg\,min}

\graphicspath{{figures/}}

\crefname{algocf}{Alg.}{Algs.}
\Crefname{algocf}{Algorithm}{Algorithms}
\crefname{section}{Sec.}{Secs.}
\Crefname{section}{Section}{Sections}
\crefname{table}{Tab.}{Tabs.}
\Crefname{table}{Table}{Tables}

\definecolor{gblue}{HTML}{4285F4}
\definecolor{gred}{HTML}{DB4437}
\definecolor{ggreen}{HTML}{0F9D58}

\acrodef{vqa}[VQA]{Visual Question Answering}
\acrodef{rpm}[RPM]{Raven's Progressive Matrices}
\acrodef{wren}[WReN]{Wild Relational Network}
\acrodef{ai}[AI]{Artificial Intelligence}
\acrodef{jsd}[JSD]{Jensen–Shannon Divergence}
\acrodef{alans}[ALANS]{ALgebra-Aware Neuro-Semi-Symbolic}
\acrodef{ni}[NI]{Neural Interpreter}
\acrodef{iid}[I.I.D.]{Independent and Identically Distributed}

\begin{document}
% \linenumbers
\pagestyle{headings}
\mainmatter
\def\ECCVSubNumber{0000}  % Insert your submission number here

\title{Learning Algebraic Representation for\\Systematic Generalization in Abstract Reasoning}% Replace with your title

\begin{comment}
\titlerunning{ECCV-22 submission ID \ECCVSubNumber} 
\authorrunning{ECCV-22 submission ID \ECCVSubNumber} 
\author{Anonymous ECCV submission}
\institute{Paper ID \ECCVSubNumber}
\end{comment}
%******************

% \begin{comment}
\titlerunning{Learning Algebraic Representation in Abstract Reasoning}
\author{Chi Zhang$^\star$\inst{1}\orcidlink{0000-0003-4948-0714}\index{Zhang, Chi} \and
Sirui Xie$^\star$\inst{1}\orcidlink{0000-0002-0295-2588}\index{Xie, Sirui} \and
Baoxiong Jia$^\star$\inst{1}\orcidlink{0000-0002-4968-3290}\index{Jia, Baoxiong} \\
Ying Nian Wu\inst{1}\orcidlink{0000-0001-8029-3664}\index{Wu, Ying Nian} \and
Song-Chun Zhu\inst{1,2,3,4}\orcidlink{0000-0002-1925-5973}\index{Zhu, Song-Chun} \and
Yixin Zhu\inst{2}\orcidlink{0000-0001-7024-1545}\index{Zhu, Yixin}}

\authorrunning{C. Zhang et al.}
\institute{University of California, Los Angeles, CA 90095, USA \and
Institute for Artificial Intelligence, Peking University, Beijing 10080, China \and 
Tsinghua University, Beijing 10080, China \and 
Beijing Institute for General Artificial Intelligence, Beijing 10080, China\\
$^\star$ indicates equal contribution\\
\email{\{chi.zhang,srxie,baoxiongjia\}@ucla.edu, \{ywu,sczhu\}@stat.ucla.edu, yixin.zhu@pku.edu.cn}\\
Project Website: \url{http://wellyzhang.github.io/project/alans.html}}
% \end{comment}
%******************
\maketitle

\begin{abstract}
Is intelligence realized by connectionist or classicist? While connectionist approaches have achieved superhuman performance, there has been growing evidence that such task-specific superiority is particularly fragile in \emph{systematic generalization}. This observation lies in the central debate between connectionist and classicist, wherein the latter continually advocates an \emph{algebraic} treatment in cognitive architectures. In this work, we follow the classicist's call and propose a hybrid approach to improve systematic generalization in reasoning. Specifically, we showcase a prototype with algebraic representation for the abstract spatial-temporal reasoning task of \ac{rpm} and present the \ac{alans} learner. The \ac{alans} learner is motivated by abstract algebra and the representation theory. It consists of a neural visual perception frontend and an algebraic abstract reasoning backend: the frontend summarizes the visual information from object-based representation, while the backend transforms it into an algebraic structure and induces the hidden operator on the fly. The induced operator is later executed to predict the answer's representation, and the choice most similar to the prediction is selected as the solution. Extensive experiments show that by incorporating an algebraic treatment, the \ac{alans} learner outperforms various pure connectionist models in domains requiring systematic generalization. We further show the generative nature of the learned algebraic representation; it can be decoded by isomorphism to generate an answer.
\end{abstract}

%%%%%%%%% BODY TEXT
\section{Introduction}

\begin{quote}
``Thought is in fact a kind of Algebra.''

\hfill---William James~\cite{james1891principles}
\end{quote}

Imagine you are given two alphabetical sequences of ``$c, b, a$'' and ``$d, c, b$'', and asked to fill in the missing element in ``$e, d, ?$''. In nearly no time will one realize the answer to be $c$. However, more surprising for human learning is that, effortlessly and instantaneously, we can ``freely generalize''~\cite{marcus2001algebraic} the solution to any partial consecutive ordered sequences. While believed to be innate in early development for human infants~\cite{marcus1999rule}, such systematic generalizability has constantly been missing and proven to be particularly challenging in existing connectionist models~\cite{bahdanau2018systematic,lake2018generalization}. In fact, such an ability to entertain a given thought and semantically related contents strongly implies an abstract algebra-like treatment~\cite{fodor1988connectionism}; in literature, it is referred to as the ``language of thought''~\cite{fodor1975language}, ``physical symbol system''~\cite{newell1980physical}, and ``algebraic mind''~\cite{marcus2001algebraic}. However, in stark contrast, existing connectionist models tend only to capture statistical correlation~\cite{chollet2019measure,kansky2017schema,lake2018generalization}, rather than providing any account for a structural inductive bias where systematic algebra can be carried out to facilitate generalization.

This contrast instinctively raises a question---what constitutes such an \emph{algebraic} inductive bias? We argue that the foundation of modeling counterpart to the algebraic treatment in early human development~\cite{marcus2001algebraic,marcus1999rule} lies in algebraic computations set up on mathematical axioms, a form of formalized human intuition and the beginning of modern mathematical reasoning~\cite{heath1956thirteen,maddy1988believing}. Of particular importance to the algebra's basic building blocks is the Peano Axiom~\cite{peano1889arithmetices}. In the Peano Axiom, the essential components of algebra, the algebraic set, and corresponding operators over it, are governed by three statements: (1) the existence of at least one element in the field to study (``zero'' element), (2) a successor function that is recursively applied to all elements and can, therefore, span the entire field, and (3) the principle of mathematical induction. Building on such a fundamental axiom, we begin to form the notion of an algebraic set and induce the operator to construct an algebraic structure. We hypothesize that such an algebraic treatment set up on fundamental axioms is essential for a model's systematic generalizability, the lack of which will only make it sub-optimal.

To demonstrate the benefits of adopting such an algebraic treatment in systematic generalization, we showcase a prototype for \acf{rpm}~\cite{raven1936mental,raven1998raven}, an exemplar task for abstract spatial-temporal reasoning~\cite{santoro2018measuring,zhang2019raven}. In this task, an agent is given an incomplete $3\times3$ matrix consisting of eight context panels with the last one missing, and asked to pick one answer from a set of eight choices that best completes the matrix. Human's reasoning capability of solving this abstract reasoning task has been commonly regarded as an indicator of ``general intelligence''~\cite{carpenter1990one} and ``fluid intelligence''~\cite{hofstadter1995fluid,jaeggi2008improving,spearman1923nature,spearman1927abilities}. In spite of the task being one that ideally requires abstraction, algebraization, induction, and generalization~\cite{carpenter1990one,raven1936mental,raven1998raven}, recent endeavors unanimously propose pure connectionist models that attempt to circumvent such intrinsic cognitive requirements~\cite{hu2020hierarchical,santoro2018measuring,wang2020abstract,wu2020scattering,zhang2019raven,zhang2019learning,zheng2019abstract}. However, these methods' inefficiency is also evident in systematic generalization; they struggle to extrapolate to domains beyond training~\cite{santoro2018measuring,zhang2019learning}, shown also in this paper.

To address the issue, we introduce an \acf{alans} learner. At a high-level, the \ac{alans} learner is embedded in a general neuro-symbolic architecture~\cite{han2019visual,mao2019neuro,yi2020clevrer,yi2018neural} but has the on-the-fly operator learnability, hence \textbf{semi-symbolic}. Specifically, it consists of a neural visual perception frontend and an algebraic abstract reasoning backend. For each \ac{rpm} instance, the neural visual perception frontend first slides a window over each panel to obtain the object-based representation~\cite{kansky2017schema,wu2017neural} for every object. A belief inference engine latter aggregates all object-based representation in each panel to produce the probabilistic \emph{belief state}. The algebraic abstract reasoning backend then takes the belief states of the eight context panels, treats them as snapshots on an algebraic structure, lifts them into a matrix-based algebraic representation built on the Peano Axiom and the representation theory~\cite{humphreys2012introduction}, and induces the hidden operator in the algebraic structure by solving an inner optimization problem~\cite{bard2013practical,colson2007overview,zhang2019metastyle}. The answer's algebraic representation is predicted by executing the induced operator: its corresponding set element is decoded by isomorphism, and the final answer is selected as the one most similar to the prediction.

The \ac{alans} learner enjoys several benefits in abstract reasoning with an algebraic treatment:
\begin{enumerate}
    \item Unlike previous monolithic models, the \ac{alans} learner offers a more \textbf{interpretable} account of the entire abstract reasoning process: the neural visual perception frontend extracts object-based representation and produces belief states of panels by explicit probability inference, whereas the algebraic abstract reasoning backend induces the hidden operator in the algebraic structure. The final answer's representation is obtained by executing the induced operator, and the choice panel with minimum distance is selected. This process much resembles the top-down bottom-up strategy in human reasoning missed in recent literature~\cite{hu2020hierarchical,santoro2018measuring,wang2020abstract,wu2020scattering,zhang2019raven,zhang2019learning,zheng2019abstract}: humans reason by inducing the hidden relation, executing it to generate a feasible solution in mind, and choosing the most similar answer available~\cite{carpenter1990one}.
    
    \item While keeping the semantic interpretability and end-to-end trainability in existing neuro-symbolic frameworks~\cite{han2019visual,mao2019neuro,yi2020clevrer,yi2018neural}, \ac{alans} is \textbf{semi-symbolic} in the sense that the symbolic operator can be learned and concluded on the fly without manual definition for every one of them. Such an inductive ability also enables a greater extent of the desired generalizability.
    
    \item By decoding the predicted representation in the algebraic structure, we can also generate an answer that satisfies the hidden relation in the context.
\end{enumerate}

This work makes three major contributions. (1) We propose the \ac{alans} learner, a neuro-semi-symbolic design, in contrast to existing monolithic models. (2) To demonstrate the efficacy of incorporating an algebraic treatment in reasoning, we show the superior systematic generalization ability of the proposed \ac{alans} learner in various extrapolatory \ac{rpm} domains. (3) We present analyses into both neural visual perception and algebraic abstract reasoning. 

\section{Related Work}\label{sec:related}

\subsection{Quest for Symbolized Manipulation}

The idea to treat thinking as a mental language can be dated back to Augustine~\cite{augustine1876confessions,wittgenstein1953philosophical}. Since the 1970s, this school of thought has undergone a dramatic revival as the quest for symbolized manipulation in cognitive modeling, such as ``language of thought''~\cite{fodor1975language}, ``physical symbol system''~\cite{newell1980physical}, and ``algebraic mind''~\cite{marcus2001algebraic}. In their study, connectionist's task-specific superiority and inability to generalize beyond training~\cite{chollet2019measure,kansky2017schema,santoro2018measuring,zhang2019raven} have been hypothetically linked to a lack of such symbolized algebraic manipulation~\cite{chollet2019measure,lake2018generalization,marcus2020next}. With evidence that an algebraic treatment adopted in early human development~\cite{marcus1999rule} can potentially address the issue~\cite{bahdanau2018systematic,mao2019neuro,marcus2020next}, classicist~\cite{fodor1988connectionism} approaches for generalizable reasoning used in programs~\cite{mccarthy1960programs} and blocks world~\cite{winograd1971procedures} have resurrected. As a hybrid approach to bridge connectionist and classicist, recent developments lead to neuro-symbolic architectures. In particular, the community of theorem proving has been one of the earliest to endorse the technique~\cite{garcez2012neural,rocktaschel2017end,serafini2016logic}: $\partial$ILP~\cite{evans2018learning} and NLM~\cite{dong2018neural} make inductive programming end-to-end, and DeepProbLog~\cite{manhaeve2018deepproblog} connects learning and reasoning. Recently, Hudson and Manning~\cite{hudson2019learning} propose NSM for visual question answering where a probabilistic graph is used for reasoning. Yi \etal~\cite{yi2018neural} demonstrate a neuro-symbolic prototype for the same task where a perception module and a language parsing module are separately trained, with the predefined logic operators associated with language tokens chained to process the visual information. Mao \etal~\cite{mao2019neuro} soften the predefined operators to afford end-to-end training with only question answers. Han \etal~\cite{han2019visual} use the hybrid architecture for metaconcept learning. Yi \etal~\cite{yi2020clevrer} and Chen \etal~\cite{chen2020grounding} show how neuro-symbolic models can handle explanatory, predictive, and counterfactual questions in temporal and causal reasoning. Lately, NeSS~\cite{chen2020compositional} exemplifies an algorithmic stack machine that can be used to improve generalization in language learning. \ac{alans} follows the classicist's call but adopts a neuro-\emph{semi}-symbolic architecture: it is end-to-end trainable as opposed to Yi \etal~\cite{yi2020clevrer,yi2018neural} and the operator can be learned and concluded on the fly without manual specification.

\subsection{Abstract Visual Reasoning}

Recent works by Santoro \etal~\cite{santoro2018measuring} and Zhang \etal~\cite{zhang2019raven} arouse the community's interest in abstract visual reasoning; the task of \acf{rpm} is introduced as such a measure for intelligent agents. As an intelligence quotient test for humans~\cite{raven1936mental,raven1998raven}, \ac{rpm} is believed to be strongly correlated with human's general intelligence~\cite{carpenter1990one} and fluid intelligence~\cite{hofstadter1995fluid,jaeggi2008improving,spearman1923nature,spearman1927abilities}. Early \ac{rpm}-solving systems employ symbolic representation based on hand-designed features and assume access to the underlying logics~\cite{carpenter1990one,lovett2017modeling,lovett2010structure,lovett2009solving}. Another stream of research on \ac{rpm} recruits similarity-based metrics to select the most similar answer from the choices~\cite{holyoak2022semantic,little2012bayesian,mcgreggor2014confident,mcgreggor2014fractals,mekik2018similarity,shegheva2018structural}. However, these visual or semantic features are unable to handle uncertainty from imperfect perception, and directly assuming access to the logic operations simplifies the problem. Recently proposed data-driven approaches arise from the availability of large datasets: Santoro \etal~\cite{santoro2018measuring} extend a pedagogical \ac{rpm} generation method~\cite{wang2015automatic}, whereas Zhang \etal~\cite{zhang2019raven} use a stochastic image grammar~\cite{zhu2007stochastic} and introduce structural annotations in it, which Hu \etal~\cite{hu2020hierarchical} further refine to avoid shortcut solutions by statistics in candidate panels. Despite the fact that \ac{rpm} intrinsically requires one to perform abstraction, algebraization, induction, and generalization, existing methods bypass such cognitive requirements using a single feedforward pass in connectionist models: Santoro \etal~\cite{santoro2018measuring} use a relational module~\cite{santoro2017simple}, Steenbrugge \etal~\cite{steenbrugge2018improving} augment it with a VAE~\cite{kingma2013auto}, Zhang \etal~\cite{zhang2019raven} assemble a dynamic tree, Hill \etal~\cite{hill2019learning} arrange the data in a contrastive manner, Zhang \etal~\cite{zhang2019learning} propose a contrast module, Zhang \etal~\cite{zheng2019abstract} formulate it in a student-teacher setting, Wang \etal~\cite{wang2020abstract} build a multiplex graph network, Hu \etal~\cite{hu2020hierarchical} aggregate features from a hierarchical decomposition, and Wu \etal~\cite{wu2020scattering} apply a scattering transformation to learn objects, attributes, and relations. Recently, Zhang \etal~\cite{zhang2021abstract} employ a neuro-symbolic design but requires full knowledge over the hidden relations to perform \emph{abduction}. While our work adopts the visual perception module and employs a similar training strategy from Zhang \etal~\cite{zhang2021abstract}, the \ac{alans} learner manages to \emph{induce} the hidden relations, enabling on-the-fly relation induction and systematic generalization on relational learning. The recent work of \ac{ni}~\cite{rahaman2021dynamic} is a complementary neural approach to our method: Although both \ac{ni} and \ac{alans} decompose the reasoning process into sub-components and aggregate them, \ac{ni} focuses more on compositionality, routing new input via different paths of learned modules to generalize, whereas \ac{alans} more on induction, enabling a learned module to adapt on the fly.

\section{The \texorpdfstring{\ac{alans}}{} Learner}\label{sec:method}

\begin{figure*}[t!]
    \centering
    \includegraphics[width=.9\linewidth]{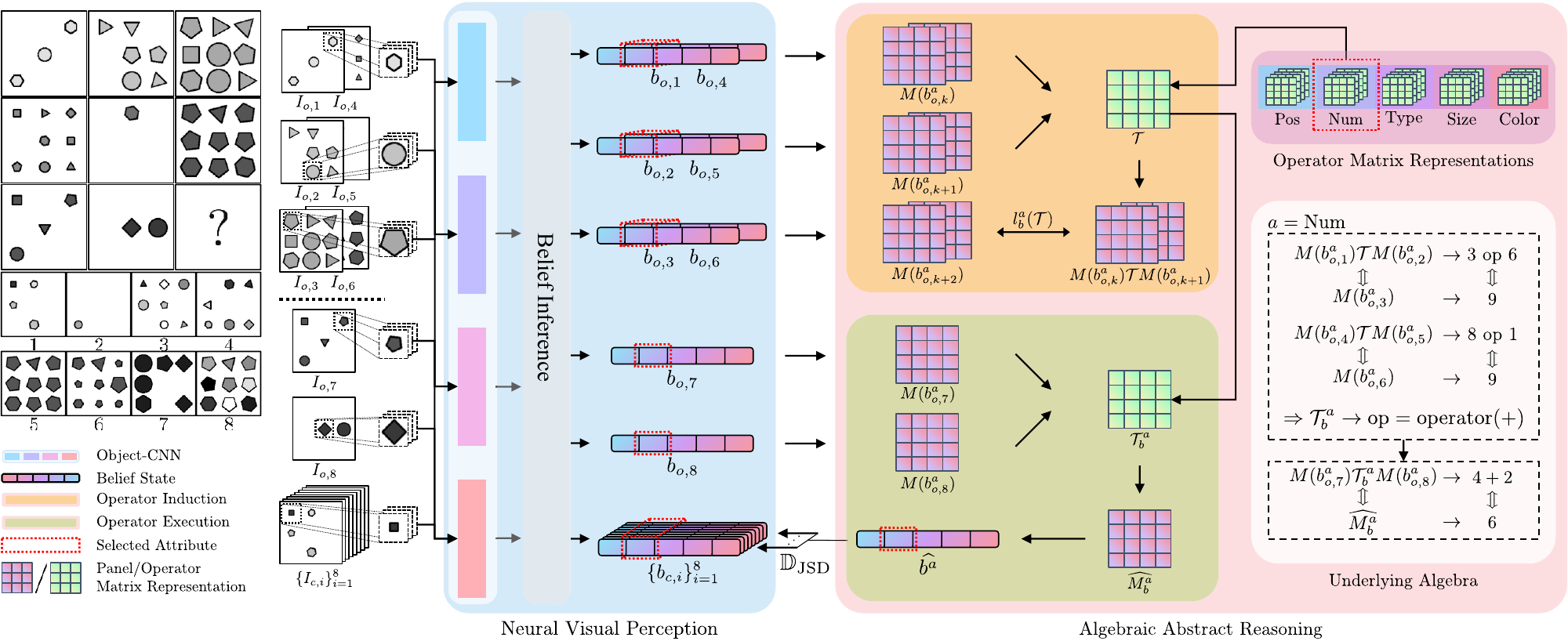}
    \caption{\textbf{An overview of the \ac{alans} learner.} For an \ac{rpm} instance, the neural visual perception module produces the belief states for all panels: an object CNN extracts object attribute distributions for each image region, and a belief inference engine marginalizes them out to obtain panel attribute distributions. For each panel attribute, the algebraic abstract reasoning module transforms the belief states into matrix-based algebraic representation and induces hidden operators by solving inner optimizations. The answer representation is obtained by executing the induced operators, and the choice most similar to the prediction is selected as the solution. An example of the underlying discrete algebra and its correspondence is also shown on the right.}
    \label{fig:overview}
\end{figure*}

In this section, we introduce the \ac{alans} learner for the \ac{rpm} problem. In each \ac{rpm} instance, an agent is given an incomplete $3\times3$ panel matrix with the last entry missing and asked to induce the operator hidden in the matrix and choose from eight choice panels one that follows it. Formally, let the answer variable be denoted as $y$, the context panels as $\{I_{o, i}\}_{i=1}^8$, and choice panels as $\{I_{c, i}\}_{i=1}^8$. Then the problem can be formulated as estimating $P(y \mid \{I_{o, i}\}_{i=1}^8, \{I_{c, i}\}_{i=1}^8)$. According to the common design~\cite{carpenter1990one,santoro2018measuring,zhang2019raven}, there is one operator that governs each panel attribute. Hence, by assuming independence among attributes, we propose to factorize the probability of $P(y = n \mid \{I_{o, i}\}_{i=1}^8, \{I_{c, i}\}_{i=1}^8)$ as
\begin{equation}
    \prod_a \sum_{\mathcal{T}^a} P(y^a = n \mid \mathcal{T}^a, \{I_{o, i}\}_{i=1}^8, \{I_{c, i}\}_{i=1}^8) \times P(\mathcal{T}^a \mid \{I_{o, i}\}_{i=1}^8),
    \label{eqn:factor}
\end{equation}
where $y^a$ denotes the answer selection based only on attribute $a$ and $\mathcal{T}^a$ the operator on $a$.

\paragraph{Overview}

As shown in \cref{fig:overview}, the \ac{alans} learner decomposes the process into perception and reasoning: the neural visual perception frontend is adopted from Zhang \etal~\cite{zhang2021abstract} and extracts the \emph{belief states} from each of the sixteen panels, whereas the algebraic abstract reasoning backend views an instance as an example in an abstract algebra structure, transforms belief states into \emph{algebraic representation} by the representation theory, \emph{induces} the hidden operators, and \emph{executes} the operators to predict the representation of the answer. Therefore, in \cref{eqn:factor}, the operator distribution is modeled by the fitness of an operator and the answer distribution by the distance between the predicted representation and that of a candidate.

\subsection{Neural Visual Perception}

We follow the design in Zhang \etal~\cite{zhang2021abstract} and decompose visual perception into an object CNN and a belief state inference engine. Specifically, for each panel, we use a sliding window to traverse the spatial domain of the image and feed each image region into an object CNN. The CNN has four branches, producing for each region its object attribute distributions, including objectiveness (if the region contains an object), type, size, and color. The belief inference engine summarizes the panel attribute distributions (over position, number, type, size, and color) by marginalizing out all object attribute distributions (over objectiveness, type, size, and color). As an example, the distribution of the panel attribute of Number can be computed as such: for $N$ image regions and their predicted objectiveness
\begin{equation}
    P(\text{Number} = k) = \underset{\substack{R^o \in \{0, 1\}^N \\ \sum_j R^o_j = k}}{\sum} \prod_{j = 1}^N P(r^o_j = R^o_j),
\end{equation}
where $P(r^o_j)$ denotes the $j$th region's estimated objectiveness distribution, and $R^o$ is a binary sequence of length $N$ that sums to $k$. All panel attribute distributions compose the \emph{belief state} of a panel. In the following, we denote the belief state as $b$ and the distribution of an attribute $a$ as $P(b^a)$. For more details, please refer to Zhang \etal~\cite{zhang2021abstract} and the Appendix.

\subsection{Algebraic Abstract Reasoning}

Given the belief states of both context and choice panels, the algebraic abstract reasoning backend concerns the induction of hidden operators and the prediction of answer representation for each attribute. The fitness of induced operators is used for estimating the operator distribution and the difference between the prediction and the choice panel for estimating the answer distribution.

\paragraph{Algebraic Underpinning}

Without loss of generality, here we assume row-wise operators. For each attribute, under perfect perception, the first two rows in an \ac{rpm} instance provide snapshots into an example of \emph{group}~\cite{hausmann1937theory} constrained to an integer-indexed set, a simple algebra structure that is closed under a binary operator. To see this, note that an accurate perception module would see each panel attribute as a deterministic set element. Therefore, \ac{rpm} instances with unary operators, such as progression, are group examples with special binary operators where one operand is constant. Instances with binary operators, such as arithmetics, directly follow the group properties. Those with ternary operators are ones defined on a three-tuple set from rows.

\paragraph{Algebraic Representation}

\begin{wrapfigure}{Rt!}{0.45\linewidth}
\centering
    \includegraphics[width=.9\linewidth]{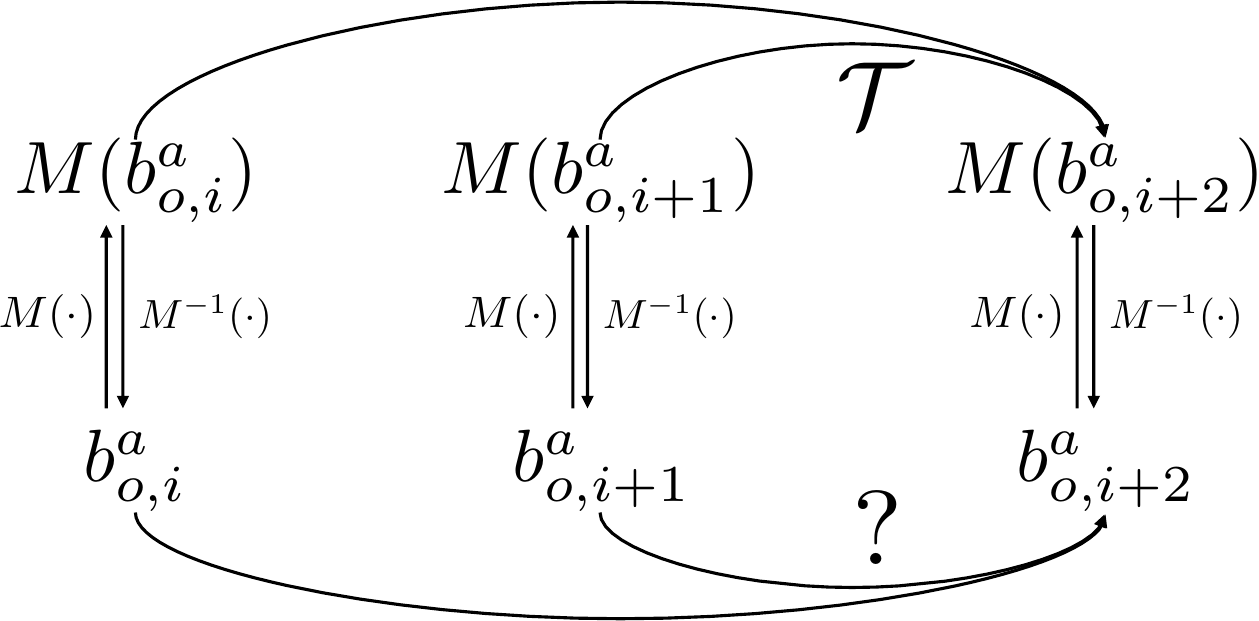}
    \caption{\textbf{Isomorphism between the abstract algebra and the matrix-based representation.} In this view, operator induction is now reduced to solving for a matrix.}
    \label{fig:isomorphism}
\end{wrapfigure}

A systematic algebraic view allows us to felicitously recruit ideas in the representation theory~\cite{humphreys2012introduction} to glean the hidden properties in the abstract structures: it makes abstract algebra amenable by reducing it onto linear algebra. Following the same spirit, we propose to lift both the set elements and the hidden operators to a learnable matrix space. To encode the set element, we employ the Peano Axiom~\cite{peano1889arithmetices}. According to the Peano Axiom, an integer-indexed set can be constructed by (1) a zero element ($\mathbf{0}$), (2) a successor function ($S(\cdot)$), and (3) the principle of mathematical induction, such that the $k$th element is encoded as $S^k(\mathbf{0})$. Specifically, we instantiate the zero element as a learnable matrix $M_0$ and the successor function as the matrix-matrix product parameterized by $M$. In an attribute-specific manner, the representation of an attribute taking the $k$th value is $(M^a)^k M^a_0$. For operators, we consider them to live in a learnable matrix group of a corresponding dimension, such that the action of an operator on a set can be represented as matrix multiplication. Such algebraic representation establishes an isomorphism between the matrix space and the abstract algebraic structure: abstract elements on the algebraic structure have a bijective mapping to/from the matrix space, and inducing the abstract relation can be reduced to solving for a matrix operator. See \cref{fig:isomorphism} for a graphical illustration of the isomorphism.

\paragraph{Operator Induction}

Operator induction concerns about finding a concrete operator in the abstract algebraic structure. By the property of closure, we formulate it as an inner-level regularized linear regression problem: a binary operator $\mathcal{T}^a_b$ for attribute $a$ in a group minimizes $\ell^a_b(\mathcal{T})$ defined as
\begin{equation}
    \ell^a_b(\mathcal{T}) = \sum_i \mathbb{E} \left[\Vert M(b^a_{o, i}) \mathcal{T} M(b^a_{o, i + 1}) - M(b^a_{o, i + 2}) \Vert^2_F \right] + \lambda^a_b \Vert \mathcal{T} \Vert^2_F,
    \label{eqn:binary}
\end{equation}
where under visual uncertainty, we take the expectation \wrt the distributions in the belief states of context panels $P(b^a_{o, i})$ in the first two rows, and denote its algebraic representation as $M(b^a_{o, i})$. For unary operators, one operand can be treated as constant and absorbed into $\mathcal{T}$. Note that \cref{eqn:binary} admits a closed-form solution (please refer to the Appendix for details). Therefore, the operator can be learned and adapted for different instances of binary relations and concluded on the fly. Such a design also simplifies the recent neuro-symbolic approaches, where every single symbol operator needs to be hand-defined~\cite{han2019visual,mao2019neuro,yi2020clevrer,yi2018neural}. Instead, we only specify an inner-level optimization framework and allow symbolic operators to be quickly induced based on the neural observations, while keeping the semantic interpretability in the neuro-symbolic methods. Therefore, we term such a design semi-symbolic.

The operator probability in \cref{eqn:factor} is then modeled by each operator type's fitness, \eg, for binary, 
\begin{equation}
    P(\mathcal{T}^a = \mathcal{T}^a_b \mid \{I_{o, i}\}^8_{i = 1})\, \propto\, \exp(-\ell^a_b(\mathcal{T}^a_b)).
    \label{eqn:operator}
\end{equation}

\paragraph{Operator Execution}

To predict the algebraic representation of the answer, we solve another inner-level optimization similar to \cref{eqn:binary}, but now treating the representation of the answer as a variable:
\begin{equation}
    \widehat{M^a_b} = \argmin_M \ell^a_b(M) = \mathbb{E}[\Vert M(b^a_{o, 7}) \mathcal{T}^a_b M(b^a_{o, 8}) - M \Vert^2_F],
\end{equation}
where the expectation is taken \wrt context panels in the last row. The optimization also admits a closed-form solution (please refer to the Appendix for details), which corresponds to the execution of the induced operator in \cref{eqn:binary}.

The predicted representation is decoded probabilistically as the predicted belief state of the solution,
\begin{equation}
    P(\widehat{b^a} = k \mid \mathcal{T}^a)\, \propto\, \exp(-\Vert \widehat{M^a} - (M^a)^k M^a_0 \Vert^2_F).
    \label{eqn:generate}
\end{equation}

\paragraph{Answer Selection}

Based on \cref{eqn:factor,eqn:operator}, estimating the answer distribution is now boiled down to estimating the conditional answer distributions for each attribute. Here, we propose to model it based on the \ac{jsd} of the predicted belief state and that of a choice,
\begin{equation}
    P(y^a = n \mid \mathcal{T}^a, \{I_{o, i}\}_{i=1}^8, \{I_{c, i}\}_{i=1}^8)\, \propto\, \exp(-d_n^a),
\end{equation}
where we define $d_n^a$ as
\begin{equation}
    d_n^a = \mathbb{D}_\text{JSD}(P(\widehat{b^a} \mid \mathcal{T}^a) \Vert P(b^a_{c, n}))).
\end{equation}

\paragraph{Discussion}

Comparing with the possible problem-solving process by humans~\cite{carpenter1990one}, we argue that the proposed algebraic abstract reasoning module offers a computational and interpretable counterpart to human-like reasoning in \ac{rpm}. Specifically, the induction component resembles fluid intelligence, where one quickly induces the hidden operator by observing the context panels. The execution component synthesizes an image by executing the induced operator, and the choice most similar to the image is selected as the answer. 

We also note that by decoding the predicted representation in \cref{eqn:generate}, a solution can be \emph{generated}: by sequentially selecting the most probable operator and the most probable attribute value, a rendering engine can directly render the solution. The reasoning backend also enables end-to-end training: by integrating the belief states from neural perception, the module conducts both induction and execution in a soft manner, such that the gradients can be back-propagated and both the visual frontend and the reasoning backend jointly trained. 

\subsection{Training Strategy}

We train the entire \ac{alans} learner by minimizing the cross-entropy loss between the estimated answer distribution and the ground-truth selection and an auxiliary loss~\cite{santoro2018measuring,wang2020abstract,zhang2019raven,zhang2021abstract} that shapes the operator distribution from the reasoning engine, \ie,
\begin{equation}
    \min_{\theta, \{M^a_0\}, \{M^a\}} \ell(P(y \mid \{I_{o, i}\}_{i=1}^8, \{I_{c, i}\}_{i=1}^8), y_\star) + \sum_a \lambda^a \ell(P(\mathcal{T}^a \mid \{I_{o, i}\}_{i=1}^8), y^a_\star),
    \label{eqn:opt}
\end{equation}
where $\ell(\cdot)$ denotes the cross-entropy loss, $y_\star$ the correct choice in candidates, and $y^a_\star$ the ground-truth operator selection for attribute $a$. The first part of the loss encourages the model to select the right choice for evaluation, while the second part motivates meaningful internal representation to emerge. Compared to Zhang \etal~\cite{zhang2021abstract}, the system requires joint operation from not only a trained perception module $\theta$, but also the algebraic encodings from the zero elements $\{M^a_0\}$ and the successor functions $\{M^a\}$, and correspondingly, induced operators $\mathcal{T}$. We notice the three-stage curriculum in Zhang \etal~\cite{zhang2021abstract} is crucial for such a neuro-semi-symbolic system. In particular, we use $\lambda^a$ to balance the trade-off in the curriculum: in the first stage, we only train parameters regarding objectiveness; in the second stage, we freeze objectiveness parameters and cyclically train parameters involving type, size, and color; in the last stage, we fine-tune all parameters.

\section{Experiments}\label{sec:experiments}

A cognitive architecture with systematic generalization is believed to demonstrate the following three principles~\cite{fodor1988connectionism,marcus2001algebraic,marcus2020next}: (1) systematicity, (2) productivity, and (3) localism. Systematicity requires an architecture to be able to entertain ``semantically related'' contents after understanding a given thought. Productivity states the awareness of a constituent implies that of a recursive application of the constituent; vice versa for localism. 

To verify the effectiveness of an algebraic treatment in systematic generalization, we showcase the superiority of the proposed \ac{alans} learner on the three principles in the abstract spatial-temporal reasoning task of \ac{rpm}. Specifically, we use the generation methods proposed in Zhang \etal~\cite{zhang2019raven} and Hu \etal~\cite{hu2020hierarchical} to generate \ac{rpm} problems and carefully split training and testing to construct the three regimes. The former generates candidates by perturbing only one attribute of the correct answer while the later modifies attribute values in a hierarchical manner to avoid shortcut solutions by pure statistics. Both methods categorize relations in \ac{rpm} into three types, according to Carpenter \etal~\cite{carpenter1990one}: unary (Constant and Progression), binary (Arithmetic), and ternary (Distribution of Three), each of which comes with several instances. Grounding the principles into learning abstract relations in \ac{rpm}, we fix the configuration to be $3\times3$Grid and generate the following data splits for evaluation (please refer to the Appendix for details):
\begin{itemize}
    \item Systematicity: the training set contains only a subset of instances for each type of relation, while the test set all other relation instances.
    \item Productivity: as the binary relation results from a recursive application of the unary relation, the training set contains only unary relations, whereas the test set only binary relations.
    \item Localism: the training and testing sets in the productivity split are swapped to study localism.
\end{itemize}

We follow Zhang \etal~\cite{zhang2019raven} to generate $10,000$ instances for each split and assign $6$ folds for training, $2$ folds for validation, and $2$ folds for testing.

\subsection{Experimental Setup}\label{sec:setup}

We evaluate the systematic generalizability of the proposed \ac{alans} learner on the above three splits, and compare the \ac{alans} learner with other baselines, including ResNet~\cite{he2016deep}, ResNet+DRT~\cite{zhang2019raven}, WReN~\cite{santoro2018measuring}, CoPINet~\cite{zhang2019learning}, MXGNet~\cite{wang2020abstract}, LEN~\cite{zheng2019abstract}, HriNet~\cite{hu2020hierarchical}, and SCL~\cite{wu2020scattering}. We use either official or public implementations that reproduce the original results. All models are implemented in PyTorch~\cite{paszke2017automatic} and optimized using ADAM~\cite{kingma2014adam} on an Nvidia Titan Xp GPU. We validate trained models on validation sets and report performance on test sets. 

\subsection{Systematic Generalization}

\cref{tbl:sys} shows the performance of various models on systematic generalization, \ie, systematicity, productivity, and localism. Compared to results reported in existing works mentioned above, all pure connectionist models experience a devastating performance drop when it comes to the critical cognitive requirements on systematic generalization, indicating that pure connectionist models fail to perform abstraction, algebraization, induction, or generalization needed in solving the abstract reasoning task; instead, they seem to only take a shortcut to bypass them. In particular, MXGNet's~\cite{wang2020abstract} superiority is diminishing in systematic generalization. In spite of learning with structural annotations, ResNet+DRT~\cite{zhang2019raven} does not fare better than its base model. The recently proposed HriNet~\cite{hu2020hierarchical} slightly improves on ResNet~\cite{he2016deep} in this aspect, with LEN~\cite{zheng2019abstract} being only marginally better. WReN~\cite{santoro2018measuring}, on the other hand, shows oscillating performance across three regimes. Evaluated under systematic generation, SCL~\cite{wu2020scattering} and CoPINet~\cite{zhang2019learning} also far deviate from ``superior performance.'' These observations suggest that pure connectionist models highly likely learn from variation in visual appearance rather than the algebra underlying the problem.

\begin{table*}[t!]
    \centering
    \caption{\textbf{Model performance on different aspects of systematic generalization.} The performance is measured by accuracy on the test sets. Results on datasets generated by Zhang \etal~\cite{zhang2019raven} (upper) and by Hu \etal~\cite{hu2020hierarchical} (lower).}
    \label{tbl:sys}
    \resizebox{\linewidth}{!}{%
        \begin{tabular}{l c c c c c c c c c c c}
            \toprule
            Method        & MXGNet    & ResNet+DRT & ResNet    & HriNet    & LEN       & WReN      & SCL       & CoPINet   & \ac{alans}         & \ac{alans}-Ind & \ac{alans}-V \\
            \midrule
            Systematicity & $20.95\%$ & $33.00\%$  & $27.35\%$ & $28.05\%$ & $40.15\%$ & $35.20\%$ & $37.35\%$ & $59.30\%$ & $\mathbf{78.45\%}$ & $52.70\%$      & $93.85\%$     \\
            Productivity  & $30.40\%$ & $27.95\%$  & $27.05\%$ & $31.45\%$ & $42.30\%$ & $56.95\%$ & $51.10\%$ & $60.00\%$ & $\mathbf{79.95\%}$ & $36.45\%$      & $90.20\%$     \\
            Localism      & $28.80\%$ & $24.90\%$  & $23.05\%$ & $29.70\%$ & $39.65\%$ & $38.70\%$ & $47.75\%$ & $60.10\%$ & $\mathbf{80.50\%}$ & $59.80\%$      & $95.30\%$     \\
            \midrule
            Average       & $26.72\%$ & $28.62\%$  & $25.82\%$ & $29.73\%$ & $40.70\%$ & $43.62\%$ & $45.40\%$ & $59.80\%$ & $\mathbf{79.63\%}$ & $48.65\%$      & $93.12\%$     \\
            \midrule
            Systematicity & $13.35\%$ & $13.50\%$  & $14.20\%$ & $21.00\%$ & $17.40\%$ & $15.00\%$ & $24.90\%$ & $18.35\%$ & $\mathbf{64.80\%}$ & $52.80\%$      & $84.85\%$     \\
            Productivity  & $14.10\%$ & $16.10\%$  & $20.70\%$ & $20.35\%$ & $19.70\%$ & $17.95\%$ & $22.20\%$ & $29.10\%$ & $\mathbf{65.55\%}$ & $32.10\%$      & $86.55\%$     \\
            Localism      & $15.80\%$ & $13.85\%$  & $17.45\%$ & $24.60\%$ & $20.15\%$ & $19.70\%$ & $29.95\%$ & $31.85\%$ & $\mathbf{65.90\%}$ & $50.70\%$      & $90.95\%$     \\
            \midrule
            Average       & $14.42\%$ & $14.48\%$  & $17.45\%$ & $21.98\%$ & $19.08\%$ & $17.55\%$ & $25.68\%$ & $26.43\%$ & $\mathbf{65.42\%}$ & $45.20\%$      & $87.45\%$     \\
            \bottomrule
        \end{tabular}%
    }
\end{table*}

Embedded in a neural-semi-symbolic framework, the proposed \ac{alans} learner improves on systematic generalization by a large margin. With an algebra-aware design, the model is considerably stable across different principles of systematic generalization. The algebraic representation learned in relations of either a constituent or a recursive composition naturally supports productivity and localism, while semi-symbolic inner optimization further allows various instances of an operator type to be induced from the algebraic representation and boosts systematicity. The importance of the algebraic representation is made more significant in the ablation study: \ac{alans}-Ind, with algebraic representation replaced by independent encodings and the algebraic isomorphism broken, shows inferior performance. We also examine the performance of the learner with perfect visual annotations (denoted as \ac{alans}-V) to see how the proposed algebraic reasoning module works: the gap despite of accurate perception indicates space for improvement for the inductive reasoning part of the model. In the next section, we further show that the neuro-semi-symbolic decomposition in \ac{alans}'s design enables diagnostic tests into its jointly learned perception module and reasoning module. This design is in stark contrast to black-box models. 

\subsection{Analysis into Perception and Reasoning}\label{sec:analysis}

The neural-semi-symbolic design affords analyses into both perception and reasoning. To evaluate the neural perception and the algebraic reasoning modules, we extract region-based object attribute annotations from the datasets~\cite{hu2020hierarchical,zhang2019raven} and categorize all relations into three types, \ie, unary, binary, and ternary.

\cref{tbl:perception} shows the perception module's performance on the test sets in the three regimes of systematic generalization. We note that in order for the \ac{alans} learner to achieve the desired results shown in \cref{tbl:sys}, \ac{alans} learns to construct the concept of objectiveness perfectly. The model also shows fairly accurate prediction on the attributes of type and size. However, on the texture-related concept of color, \ac{alans} fails to develop a reliable notion on it. Despite that, the general prediction accuracy of the perception module is still surprising, considering that the perception module is jointly learned with ground-truth annotations on answer selections. The relatively lower accuracy on color could be attributed to its larger space compared to other attributes.

\begin{table*}[ht!]
    \centering
    \caption{\textbf{Perception accuracy of the proposed \ac{alans} learner, measured by whether the module can correctly predict an attribute's value.} Results on datasets generated by Zhang \etal~\cite{zhang2019raven} (left) and by Hu \etal~\cite{hu2020hierarchical} (right).}
    \label{tbl:perception}
    \resizebox{\linewidth}{!}{
        \begin{tabular}{l c c c c}
            \toprule
            Object Attribute & Objectiveness & Type      & Size      & Color     \\
            \midrule
            Systematicity    & $100.00\%$    & $99.95\%$ & $94.65\%$ & $71.35\%$ \\
            Productivity     & $100.00\%$    & $99.97\%$ & $98.04\%$ & $77.61\%$ \\
            Localism         & $100.00\%$    & $95.65\%$ & $98.56\%$ & $80.05\%$ \\
            \midrule
            Average          & $100.00\%$    & $98.52\%$ & $97.08\%$ & $76.34\%$ \\
            \bottomrule
        \end{tabular}
        \,
        \begin{tabular}{l c c c c}
            \toprule
            Object Attribute & Objectiveness & Type      & Size      & Color     \\
            \midrule
            Systematicity    & $100.00\%$    & $96.34\%$ & $92.36\%$ & $63.98\%$ \\
            Productivity     & $100.00\%$    & $94.28\%$ & $97.00\%$ & $69.89\%$ \\
            Localism         & $100.00\%$    & $95.80\%$ & $98.36\%$ & $60.35\%$ \\
            \midrule
            Average          & $100.00\%$    & $95.47\%$ & $95.91\%$ & $64.74\%$ \\
            \bottomrule
        \end{tabular}
    }
\end{table*}

\cref{tbl:reasoning} lists the reasoning module's performance during testing for the three aspects. Note that on position, the unary operator (shifting) and binary operator (set arithmetics) do not systematically imply each other. Hence, we do not count them as probes into productivity and localism. In general, we notice that the better the perception accuracy on one attribute, the better the performance on reasoning. However, we also note that despite the relatively accurate perception of objectiveness, type, and size, near perfect reasoning is never guaranteed. This deficiency is due to the perception uncertainty handled by expectation in \cref{eqn:binary}: in spite of correctness when we take $\argmax$, marginalizing by expectation will unavoidably introduce noise into the reasoning process. Therefore, an ideal reasoning module requires the perception frontend to be not only correct but also certain. Computationally, one can sample from the perception module and optimize \cref{eqn:opt} using REINFORCE~\cite{williams1992simple}. However, the credit assignment problem and variance in gradient estimation will further complicate training.

\begin{table*}[ht!]
    \centering
    \caption{\textbf{Reasoning accuracy of the proposed \ac{alans} learner, measured by whether the module can correctly predict the type of a relation on an attribute.} Results on datasets generated by Zhang \etal~\cite{zhang2019raven} (left) and by Hu \etal~\cite{hu2020hierarchical} (right).}
    \label{tbl:reasoning}
    \resizebox{\linewidth}{!}{
        \begin{tabular}{l c c c c c}
            \toprule
            Relation on   & Position  & Number    & Type      & Size      & Color     \\
            \midrule
            Systematicity & $72.04\%$ & $82.14\%$ & $81.50\%$ & $80.80\%$ & $40.40\%$ \\
            Productivity  & -         & $98.75\%$ & $89.50\%$ & $72.10\%$ & $33.95\%$ \\
            Localism      & -         & $74.70\%$ & $44.25\%$ & $56.40\%$ & $54.20\%$ \\
            \midrule
            Average       & $72.04\%$ & $85.20\%$ & $71.75\%$ & $69.77\%$ & $42.85\%$ \\
            \bottomrule
        \end{tabular}
        \,
        \begin{tabular}{l c c c c c}
            \toprule
            Relation on   & Position  & Number    & Type      & Size      & Color     \\
            \midrule
            Systematicity & $69.96\%$ & $80.34\%$ & $83.50\%$ & $80.85\%$ & $28.85\%$ \\
            Productivity  & -         & $99.10\%$ & $87.95\%$ & $68.50\%$ & $23.10\%$ \\
            Localism      & -         & $70.55\%$ & $36.65\%$ & $42.30\%$ & $33.20\%$ \\
            \midrule
            Average       & $69.96\%$ & $83.33\%$ & $69.37\%$ & $63.88\%$ & $28.38\%$ \\
            \bottomrule
        \end{tabular}
    }
\end{table*}

\begin{table*}[ht!]
    \centering
    \caption{\textbf{Model performance on RAVEN~\cite{zhang2019raven} (left) and I-RAVEN~\cite{hu2020hierarchical} (right) under the regular I.I.D. evaluation, measured by accuracy on the test sets.}}
    \label{tbl:iid-raven}
    \resizebox{\linewidth}{!}{
        \begin{tabular}{l c c c c c c c c}
            \toprule
            Method       & Acc                               & Center                            & 2x2Grid                           & 3x3Grid                           & L-R                      & U-D                      & O-IC                     & O-IG                                          \\
            \midrule
            WReN         & $34.0\%/21.5\%$                   & $58.4\%/24.0\%$                   & $38.9\%/25.0\%$                   & $37.7\%/20.1\%$                   & $21.6\%/19.7\%$          & $19.8\%/19.9\%$          & $38.9\%/21.3\%$          & $22.6\%/20.6\%$                   \\
            ResNet       & $53.4\%/18.4\%$                   & $52.8\%/22.6\%$                   & $41.9\%/15.5\%$                   & $44.3\%/18.1\%$                   & $58.8\%/19.0\%$          & $60.2\%/19.6\%$          & $63.2\%/17.5\%$          & $53.1\%/16.6\%$                   \\
            ResNet+DRT   & $59.6\%/20.7\%$                   & $58.1\%/24.2\%$                   & $46.5\%/18.2\%$                   & $50.4\%/19.8\%$                   & $65.8\%/22.0\%$          & $67.1\%/22.1\%$          & $69.1\%/21.0\%$          & $60.1\%/18.1\%$                   \\
            LEN          & $71.6\%/32.8\%$                   & $79.1\%/44.8\%$                   & $56.1\%/27.9\%$                   & $60.3\%/23.9\%$                   & $80.5\%/34.1\%$          & $76.4\%/34.4\%$          & $79.3\%/35.8\%$          & $69.9\%/28.5\%$                   \\
            HriNet       & $45.1\%/60.8\%$                   & $66.1\%/78.2\%$                   & $40.7\%/50.1\%$                   & $38.0\%/42.4\%$	                  & $44.9\%/70.1\%$	         & $43.2\%/70.3\%$	        & $47.2\%/68.2\%$	       & $35.8\%/46.3\%$                   \\
            MXGNet       & $84.0\%/33.1\%$                   & $94.3\%/40.7\%$                   & $60.5\%/27.9\%$                   & $64.9\%/24.7\%$                   & $96.6\%/35.8\%$          & $96.4\%/34.5\%$          & $94.1\%/36.4\%$          & $81.3\%/31.6\%$                   \\
            CoPINet      & $91.4\%/46.1\%$                   & $95.1\%/54.4\%$                   & $77.5\%/36.8\%$                   & $78.9\%/31.9\%$	                  & $\mathbf{99.1\%}/51.9\%$ & $\mathbf{99.7\%}/52.5\%$	& $\mathbf{98.5\%}/52.2\%$ & $91.4\%/42.8\%$                   \\
            \ac{alans}   & $74.4\%/78.5\%$                   & $69.1\%/72.3\%$                   & $80.2\%/79.5\%$                   & $75.0\%/72.9\%$	                  & $72.2\%/79.2\%$	         & $73.3\%/79.6\%$	        & $76.3\%/85.9\%$	       & $74.9\%/79.9\%$                   \\
            SCL          & $74.2\%/80.5\%$                   & $82.8\%/84.6\%$                   & $70.4\%/79.4\%$                   & $64.1\%/69.9\%$                   & $77.6\%/82.7\%$          & $78.4\%/82.6\%$          & $84.2\%/87.3\%$          & $62.2\%/77.2\%$                   \\
            \ac{alans}-V & $\mathbf{94.4\%}/\mathbf{93.5\%}$ & $\mathbf{98.4\%}/\mathbf{98.9\%}$ & $\mathbf{91.5\%}/\mathbf{85.0\%}$ & $\mathbf{87.0\%}/\mathbf{83.2\%}$ & $97.3\%/\mathbf{90.9\%}$ & $96.4\%/\mathbf{98.1\%}$ & $97.3\%/\mathbf{99.1\%}$ & $\mathbf{93.2\%}/\mathbf{89.5\%}$ \\
            \bottomrule%
        \end{tabular}%
    }%
\end{table*}%

\subsection{In-Distribution Performance}

To further evaluate how models perform under the regular \ac{iid} setup, we train the models on the original datasets generated by Zhang \etal~\cite{zhang2019raven} and Hu \etal~\cite{hu2020hierarchical} and measure the model accuracy in the test splits. We compare \ac{alans} with published baselines in \cref{tbl:sys}.

\cref{tbl:iid-raven} (left) shows the results on the RAVEN dataset~\cite{zhang2019raven}. With the jointly trained vision component, the \ac{alans} learner does not fare better than the best connectionist approaches, making it on par with SCL only. As the dataset is known to have shortcut solutions, neural approaches like MXGNet and CoPINet could potentially find it easier to solve and hence achieve much superior results in this setup. However, \ac{alans}-V, the variant with a perfect perception component reaches a level of much robustness and accuracy, attaining the best results in grid-like layouts, empirically believed to be the hardest in human evaluation~\cite{zhang2019raven}.

\cref{tbl:iid-raven} (right) shows the results on the I-RAVEN dataset~\cite{hu2020hierarchical}. Apart from \ac{alans}-V's realizing the best performance across all models, we also notice the consistency of performance of the proposed method across datasets, with or without the shortcut issues. All other methods show drastically varying performance, particularly for CoPINet and MXGNet, arguably because of the choice generation strategy that effectively prunes easy solution paths via statistics.

In summary, by analyzing the results from \cref{tbl:sys,tbl:iid-raven} together, we notice that \ac{alans} not only attains reasonable performance on the \ac{iid} setup but also generalizes systematically.

\subsection{Generative Potential}

\begin{figure*}[t!]
    \centering
    \includegraphics[width=.9\linewidth]{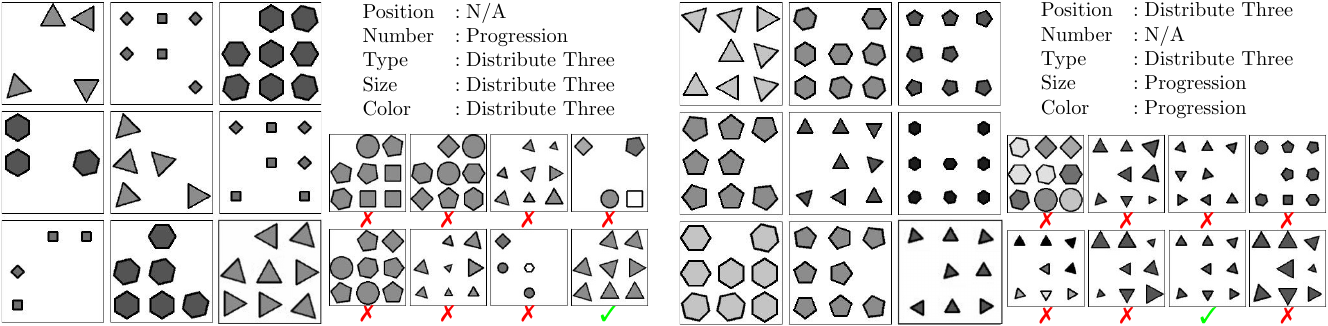}
    \caption{\textbf{Examples of \ac{rpm} instances with the missing entries filled by solutions directly generated by the \ac{alans} learner.} Ground-truth relations are also listed. Note the generated results do not look exactly the same as the correct candidate choices due to random rotations during rendering, but they are semantically correct.}
    \label{fig:generate}
\end{figure*}

Compared to existing discriminative-only \ac{rpm}-solving methods, the proposed \ac{alans} learner is unique in its generative potential. As mentioned above, the final panel attribute can be decoded by sequentially selecting the most probable hidden operator and the attribute value. A solution can be generated when equipped with a rendering engine. In \cref{fig:generate}, we use the rendering program from Zhang \etal~\cite{zhang2019raven} to showcase the generative potential in the \ac{alans} learner.

\section{Conclusion and Limitation}\label{sec:conclusion}

In this work, we propose the \acf{alans} learner, echoing a normative theory in the connectionist-classicist debate that an algebraic treatment in a cognitive architecture should improve a model's systematic generalization ability. In particular, the \ac{alans} learner employs a neural-semi-symbolic architecture, where the neural visual perception module is responsible for summarizing visual information and the algebraic abstract reasoning module transforms it into algebraic representation with isomorphism established by the Peano Axiom and the representation theory, conducts operator induction, and executes it to arrive at an answer. In three \ac{rpm} domains reflective of systematic generalization, the proposed \ac{alans} learner shows superior performance compared to other pure connectionist baselines. 

The proposed \ac{alans} learner also bears some limitations. For one thing, we make the assumption in our formulation that relations on different attributes are independent and can be factorized. This assumption is not universally correct and could potentially lead to failure in more complex reasoning scenarios when attributes are correlated. For another, we assume a fixed and known space for each attribute in the perception module, while in the real world the space for one attribute could be dynamically changing. In addition, the reasoning module is sensitive to perception uncertainty as has already been discussed in the experimental results. Besides, the gap between perfection and the status quo in reasoning remains to be filled. In this work, we only show how the hidden operator can be induced with regularized linear regression via the representation theory. However, more elaborate differentiable optimization problems can certainly be incorporated for other problems.

With the limitation in mind, we hope that this preliminary study could inspire more research on incorporating algebraic structures into current connectionist models and help address challenging modeling problems~\cite{xu2022est,zhang2021acre,zhang2020machine,zhu2020dark}.

\paragraph{Acknowledgement}

We thank Prof. Hongjing Lu and colleagues from UCLA for fruitful discussions. We would also like to thank anonymous reviewers for constructive feedback.

\bibliographystyle{splncs04}
\bibliography{bib}
\end{document}